\documentclass[runningheads]{llncs}

\RequirePackage{silence}  
\usepackage{graphicx}
\usepackage{comment}
\usepackage{amsmath,amssymb}
\usepackage{color}
\usepackage{url}
\usepackage{hyperref}

\newif\ifreview
\reviewfalse

\ifreview
	\usepackage{lineno}

	\linenumbers
\fi

\usepackage{tabularx} %
\usepackage{tikz} %
\usepackage{tikzfill} %
\usepackage{pgfplots} %
\usepackage{pgfplotstable} %
\usepackage{standalone} %
\usepackage{multirow,makecell} %
\usepackage{listings} %
\usepackage{comment} %
\usepackage{calc}  %
\usepackage{siunitx}  %
\usepackage{xspace}  %
\usepackage{accsupp} %
\usepackage{booktabs}  %

\usetikzlibrary{spy} %
\usetikzlibrary{arrows.meta} %

\pgfplotsset{compat=newest} %

\usetikzlibrary{positioning}
\usetikzlibrary{calc}
\usepgfplotslibrary{colorbrewer}
\usetikzlibrary{decorations.pathreplacing}
\usepgfplotslibrary{fillbetween}
\usetikzlibrary{patterns.meta}
\usepgfplotslibrary{groupplots}
\usetikzlibrary{plotmarks}
\usetikzlibrary{matrix}

\usepackage[capitalize]{cleveref}
\crefname{section}{Sec.}{Secs.}
\Crefname{section}{Section}{Sections}
\Crefname{table}{Table}{Tables}
\crefname{table}{Tab.}{Tabs.}

\definecolor{tabfirst}{rgb}{1, 0.7, 0.7} %
\definecolor{tabsecond}{rgb}{1, 0.85, 0.7} %

\newcommand \colorindicator[1]{%
	{\textcolor{#1}{$\blacksquare\!\!\!\!\!\blacksquare$}}%
}

\newlength{\figureWidth}

\makeatletter

\@ifclassloaded{standalone}{%
    \setlength{\figureWidth}{12.18cm}%
}{%
    \setlength{\figureWidth}{\textwidth}%
}

\pgfdeclareshape{coffeebean}{
    \inheritsavedanchors[from=circle]
    \inheritanchorborder[from=circle]
    \inheritanchor[from=circle]{center}
    \inheritanchor[from=circle]{north}
    \inheritanchor[from=circle]{south}
    \inheritanchor[from=circle]{east}
    \inheritanchor[from=circle]{west}
    \inheritanchor[from=circle]{north west}
    \inheritanchor[from=circle]{north east}
    \inheritanchor[from=circle]{south west}
    \inheritanchor[from=circle]{south east}

    \backgroundpath{
        \pgfmathsetlengthmacro{\r}{\radius} 
        \pgf@process{\centerpoint}
        \pgfmathsetlengthmacro{\cx}{\pgf@x}
        \pgfmathsetlengthmacro{\cy}{\pgf@y}
        
        \pgfpathmoveto{\pgfpoint{\cx}{\cy + \r}}
        \pgfpathcurveto
            {\pgfpoint{\cx + 0.7*\r}{\cy + \r}}
            {\pgfpoint{\cx + 0.9*\r}{\cy - \r}}
            {\pgfpoint{\cx}{\cy - \r}}
        \pgfpathcurveto
            {\pgfpoint{\cx - 0.9*\r}{\cy - \r}}
            {\pgfpoint{\cx - 0.7*\r}{\cy + \r}}
            {\pgfpoint{\cx}{\cy + \r}}
        \pgfpathclose
    }

    \foregroundpath{
        \pgfmathsetlengthmacro{\r}{\radius}
        \pgf@process{\centerpoint}
        \pgfmathsetlengthmacro{\cx}{\pgf@x}
        \pgfmathsetlengthmacro{\cy}{\pgf@y}
        
        \pgfpathmoveto{\pgfpoint{\cx}{\cy + 0.9*\r}}
        \pgfpathcurveto
            {\pgfpoint{\cx - 0.25*\r}{\cy + 0.3*\r}}
            {\pgfpoint{\cx + 0.25*\r}{\cy - 0.3*\r}}
            {\pgfpoint{\cx}{\cy - 0.9*\r}}
    }
}

\pgfdeclareshape{coffeecup}{
    \inheritsavedanchors[from=circle]
    \inheritanchorborder[from=circle]
    \inheritanchor[from=circle]{center}
    \inheritanchor[from=circle]{north}
    \inheritanchor[from=circle]{south}
    \inheritanchor[from=circle]{east}
    \inheritanchor[from=circle]{west}

    \backgroundpath{
        \pgfmathsetlengthmacro{\r}{\radius} 
        \pgf@process{\centerpoint}
        \pgfmathsetlengthmacro{\cx}{\pgf@x}
        \pgfmathsetlengthmacro{\cy}{\pgf@y}
        
        \pgfpathcircle{\pgfpoint{\cx}{\cy}}{\r}
    }

    \foregroundpath{
        \pgfmathsetlengthmacro{\r}{\radius}
        \pgf@process{\centerpoint}
        \pgfmathsetlengthmacro{\cx}{\pgf@x}
        \pgfmathsetlengthmacro{\cy}{\pgf@y}
        
        \pgfpathcircle{\pgfpoint{\cx}{\cy}}{1.15*\r}
        
        \pgfpathcircle{\pgfpoint{\cx}{\cy}}{1.6*\r}
        
        \pgfpathmoveto{\pgfpoint{\cx + 1.15*\r}{\cy + 0.4*\r}}
        \pgfpathcurveto
            {\pgfpoint{\cx + 1.8*\r}{\cy + 0.4*\r}} %
            {\pgfpoint{\cx + 1.8*\r}{\cy - 0.4*\r}} %
            {\pgfpoint{\cx + 1.15*\r}{\cy - 0.4*\r}} %
        \pgfpathcurveto
            {\pgfpoint{\cx + 1.5*\r}{\cy - 0.25*\r}} %
            {\pgfpoint{\cx + 1.5*\r}{\cy + 0.25*\r}} %
            {\pgfpoint{\cx + 1.15*\r}{\cy + 0.4*\r}} %
        \pgfpathclose
    }
}

\pgfdeclareshape{sidecup}{
    \inheritsavedanchors[from=circle]
    \inheritanchorborder[from=circle]
    \inheritanchor[from=circle]{center}
    \inheritanchor[from=circle]{north}
    \inheritanchor[from=circle]{south}
    \inheritanchor[from=circle]{east}
    \inheritanchor[from=circle]{west}

    \backgroundpath{
        \pgfmathsetlengthmacro{\r}{\radius} 
        \pgf@process{\centerpoint}
        \pgfmathsetlengthmacro{\cx}{\pgf@x}
        \pgfmathsetlengthmacro{\cy}{\pgf@y}
        
        \pgfpathmoveto{\pgfpoint{\cx - 0.9*\r}{\cy + 0.8*\r}}
        
        \pgfpathlineto{\pgfpoint{\cx - 0.7*\r}{\cy - 0.6*\r}}
        
        \pgfpathcurveto
            {\pgfpoint{\cx - 0.6*\r}{\cy - 1.1*\r}}
            {\pgfpoint{\cx + 0.6*\r}{\cy - 1.1*\r}}
            {\pgfpoint{\cx + 0.7*\r}{\cy - 0.6*\r}}
            
        \pgfpathlineto{\pgfpoint{\cx + 0.9*\r}{\cy + 0.8*\r}}
        
        \pgfpathcurveto
            {\pgfpoint{\cx + 0.9*\r}{\cy + 1.1*\r}}
            {\pgfpoint{\cx - 0.9*\r}{\cy + 1.1*\r}}
            {\pgfpoint{\cx - 0.9*\r}{\cy + 0.8*\r}}
            
        \pgfpathclose
    }

    \foregroundpath{
        \pgfmathsetlengthmacro{\r}{\radius}
        \pgf@process{\centerpoint}
        \pgfmathsetlengthmacro{\cx}{\pgf@x}
        \pgfmathsetlengthmacro{\cy}{\pgf@y}
        
        \pgfpathmoveto{\pgfpoint{\cx + 0.85*\r}{\cy + 0.4*\r}}
        \pgfpathcurveto
            {\pgfpoint{\cx + 1.6*\r}{\cy + 0.4*\r}} %
            {\pgfpoint{\cx + 1.6*\r}{\cy - 0.5*\r}} %
            {\pgfpoint{\cx + 0.73*\r}{\cy - 0.5*\r}} %
            
        \pgfpathmoveto{\pgfpoint{\cx - 0.9*\r}{\cy + 0.8*\r}}
        \pgfpathcurveto
            {\pgfpoint{\cx - 0.9*\r}{\cy + 0.5*\r}}
            {\pgfpoint{\cx + 0.9*\r}{\cy + 0.5*\r}}
            {\pgfpoint{\cx + 0.9*\r}{\cy + 0.8*\r}}

    }
}

\DeclareRobustCommand\onedot{\futurelet\@let@token\@onedot}
\def\@onedot{\ifx\@let@token.\else.\null\fi\xspace}

\def\eg{\emph{e.g}\onedot} 
\def\ie{\emph{i.e}\onedot} 
\def\cf{\emph{cf}\onedot}

\def\etal{\emph{et al}\onedot}

\makeatother

\DeclareRobustCommand{\coffeebsmall}{%
    \scalebox{0.5}{\tikz[baseline=-0.6ex]{%
        \node[coffeebean, draw=black, thick, minimum size=1em, inner sep=0pt] {};
    }}%
}

\usepackage[normalem]{ulem} %

\usepackage[table]{xcolor} %

\colorlet{best}{orange!25}
\colorlet{overallbest}{red!25}
\newcommand{\bestcell}[0]{\cellcolor{best}}
\newcommand{\overallbestcell}[0]{\cellcolor{overallbest}}

\pgfdeclareplotmark{hexagon*}
{%
  \pgfpathmoveto{\pgfqpoint{1\pgfplotmarksize}{0pt}}%
  \pgfpathlineto{\pgfqpoint{0.5\pgfplotmarksize}{0.866\pgfplotmarksize}}%
  \pgfpathlineto{\pgfqpoint{-0.5\pgfplotmarksize}{0.866\pgfplotmarksize}}%
  \pgfpathlineto{\pgfqpoint{-1\pgfplotmarksize}{0pt}}%
  \pgfpathlineto{\pgfqpoint{-0.5\pgfplotmarksize}{-0.866\pgfplotmarksize}}%
  \pgfpathlineto{\pgfqpoint{0.5\pgfplotmarksize}{-0.866\pgfplotmarksize}}%
  \pgfpathclose
  \pgfusepathqfillstroke
}

\newcommand{\MAED}{MAE\textsubscript{\coffeebsmall}\@\xspace}
\newcommand{\MAEA}{MAE\textsubscript{$\square$}\@\xspace}

\newcommand{\diff}[1]{\textcolor{darkgray!75}{\scriptsize$(#1)$}} %

\definecolor{tud0d}{RGB}{83,83,83}
\definecolor{tud0c}{RGB}{137,137,137}
\definecolor{tud0b}{RGB}{181,181,181}
\definecolor{tud0a}{RGB}{220,220,220}
\definecolor{tud1a}{RGB}{93,133,195}
\definecolor{tud2a}{RGB}{0,156,218}
\definecolor{tud3a}{RGB}{80,182,149}
\definecolor{tud4a}{RGB}{175,204,80}
\definecolor{tud5a}{RGB}{221,223,72}
\definecolor{tud6a}{RGB}{255,224,92}
\definecolor{tud7a}{RGB}{248,186,60}
\definecolor{tud8a}{RGB}{238,122,52}
\definecolor{tud9a}{RGB}{233,80,62}
\definecolor{tud10a}{RGB}{201,48,142}
\definecolor{tud11a}{RGB}{128,69,151}
\definecolor{tud1b}{RGB}{0,90,169}
\definecolor{tud2b}{RGB}{0,131,204}
\definecolor{tud3b}{RGB}{0,157,129}
\definecolor{tud4b}{RGB}{153,192,0}
\definecolor{tud5b}{RGB}{201,212,0}
\definecolor{tud6b}{RGB}{253,202,0}
\definecolor{tud7b}{RGB}{245,163,0}
\definecolor{tud8b}{RGB}{236,101,0}
\definecolor{tud9b}{RGB}{230,0,26}
\definecolor{tud10b}{RGB}{166,0,132}
\definecolor{tud11b}{RGB}{114,16,133}
\definecolor{tud1c}{RGB}{0,78,138}
\definecolor{tud2c}{RGB}{0,104,157}
\definecolor{tud3c}{RGB}{0,136,119}
\definecolor{tud4c}{RGB}{127,171,22}
\definecolor{tud5c}{RGB}{177,189,0}
\definecolor{tud6c}{RGB}{215,172,0}
\definecolor{tud7c}{RGB}{210,135,0}
\definecolor{tud8c}{RGB}{204,76,3}
\definecolor{tud9c}{RGB}{185,15,34}
\definecolor{tud10c}{RGB}{149,17,105}
\definecolor{tud11c}{RGB}{97,28,115}
\definecolor{tud1d}{RGB}{36,53,114}
\definecolor{tud2d}{RGB}{0,78,115}
\definecolor{tud3d}{RGB}{0,113,94}
\definecolor{tud4d}{RGB}{106,139,55}
\definecolor{tud5d}{RGB}{153,166,4}
\definecolor{tud6d}{RGB}{174,142,0}
\definecolor{tud7d}{RGB}{190,111,0}
\definecolor{tud8d}{RGB}{169,73,19}
\definecolor{tud9d}{RGB}{156,28,38}
\definecolor{tud10d}{RGB}{115,32,84}
\definecolor{tud11d}{RGB}{76,34,106}

\definecolor{sa1}{RGB}{80,70,130}
\definecolor{sa2}{RGB}{160,200,100}
\definecolor{sa3}{RGB}{130,30,20}

\definecolor{visinfBlue}{HTML}{005aa9}
\definecolor{ivaGreen}{HTML}{00780c}

\colorlet{pipelineW}{visinfBlue}
\colorlet{pipelineI}{ivaGreen}

\DeclareSIUnit{\px}{px}

\makeatletter
\let\titleold\title
\renewcommand{\title}[1]{\titleold{#1}\newcommand{\thetitle}{#1}}
\def\maketitlesupplementary
   {
   \newpage
        {\centering
        \Large
        \textbf{\thetitle}\\
        \vspace{0.5em}Supplementary Material \\
        \vspace{1.0em}
        }
    }

\usepackage{placeins}

\usepackage{orcidlink}
\usepackage{marvosym}

\usepackage{fancyhdr}

\begin{document}

\def\SubNumber{23}

\def\GCPRTrack{Special Track: Computer vision systems and applications}

\title{\emph{Doppio}: A Dataset for Contactless Weight Estimation of Falling Particles}

\reviewfalse

\ifreview
	\titlerunning{GCPR 2026 Submission \SubNumber{}. CONFIDENTIAL REVIEW COPY.}
	\authorrunning{GCPR 2026 Submission \SubNumber{}. CONFIDENTIAL REVIEW COPY.}
	\author{GCPR 2026 - \GCPRTrack{}}
	\institute{Paper ID \SubNumber}
\else
    \author{
        Simon Kiefhaber\inst{*,1\,}\orcidlink{0009-0000-7659-6435} \and
        Jan-Martin O. Steitz\inst{*,1\,}\orcidlink{0000-0002-3549-312X} \and
        Julia Grabinski\inst{1\,}\orcidlink{0000-0002-8371-1734} \and\\
        Christoph Reich\inst{1,2,3,4\,}\orcidlink{0000-0002-8616-1627} \and
        Paul Wagner\inst{1\,}\orcidlink{0000-0002-6125-574X} \and
        Max Zimmermann\inst{1\,}\orcidlink{0009-0002-5462-3804} \and\\
        Simone Schaub-Meyer\inst{1,3,5\,}\orcidlink{0000-0001-8644-1074} \and
        Stefan Roth\inst{1,3,5\,}\orcidlink{0000-0001-9002-9832}
    }
	
	\authorrunning{S. Kiefhaber \& J.-M. Steitz \emph{et al.}}
	
    \institute{
        \textsuperscript{1}\,TU Darmstadt\;
        \textsuperscript{2}\,TU Munich\;
        \textsuperscript{3}\,ELIZA\;
        \textsuperscript{4}\,MCML\;
        \textsuperscript{5}\,hessian.AI\;
        \textsuperscript{*}\,equal contribution\\
        \email{\{name.surname\}@visinf.tu-darmstadt.de}\\
        \url{https://visinf.github.io/doppio}
    }
\fi

\maketitle              %
\thispagestyle{fancy}  

\begin{abstract}
Measuring the mass of powder, including falling particles, is a common task in industrial applications. While scales are effective for static measurements, many applications require contactless sensing, where existing solutions are often costly, application-specific, and technically complex. In this work, we investigate computer vision as a practical alternative for contactless mass estimation. As an accessible real-world case study, we focus on coffee grinding and introduce \emph{Doppio}, a novel video dataset capturing videos of falling ground coffee, paired with precise, per-frame ground-truth weight measurements. To demonstrate contactless measuring, we evaluate deep learning-based approaches ranging from purely spatial feed-forward networks to recurrent spatio-tempo\-ral models. These models are analyzed with respect to their predictive accuracy and computational trade-offs. We demonstrate that deep learning-based computer vision models accurately estimate the cumulative weight of falling particles, establishing a solid foundation for future vision-based contactless measurement solutions.
\end{abstract}
\section{Introduction}

The precise measurement of powder mass is a fundamental task common in numerous industry applications, including solid dosage in pharmaceutical manufacturing \cite{Blackshields:2018:CPF}, additive manufacturing via laser metal deposition \cite{Breese:2021:MCC}, and the food industry \cite{Kruppa:2023:IDA}. Mechanical vibrations often prohibit the application of scales or weight cells. Additionally, they only support batch processing, restricting material flow. For a continuous mode of manufacturing or contactless applications, other specialized measurement approaches are required. These often include highly specialized hardware such as microwave radar or X-ray~\cite{Ganesh:2017:AXS,Isa:2006:MDR}, significantly increasing engineering effort, technical complexity, and total production cost. A simple, cost-efficient, and general contactless measuring solution remains lacking.

Therefore, there is a significant incentive to move toward off-the-shelf contactless measurement solutions. Computer vision offers a compelling alternative. Optical sensors are less prone to mechanical vibration and universally available as low-cost, high-resolution video cameras. Additionally, current computer vision approaches demonstrate effective video understanding~\cite{Lin:2019:TSM,Tang:2026:VID}. Video-based weight estimation using computer vision can aid manufacturers reduce costs and technical complexity while increasing system durability.

As access to industry-scale systems that use powder flow is limited, we turn to coffee bean grinding as a \emph{case study}. The grinding of coffee beans provides a relevant setting, as industry applications frequently rely on the dosing of fine-grained materials. Taking a closer look, during coffee grinding, coffee particles agglomerate mainly due to the triboelectric effect \cite{MendezHarper:2024:MCT}, but also because of capillary liquid bridges from the lipid fraction released from the coffee beans \cite{Herminghaus:2005:DWG,Speer:2006:LFC}, and powder caking due to compaction in the grinder \cite{Dogan:2019:PCC}. As a result, depending on agglomerate size, we observe different speeds and occlusions of smaller clumps and particles, making the weight estimation of falling ground coffee a challenging problem and, therefore, an adequate substitute model.

To the best of our knowledge, no published work has explored the use of computer vision for dynamic and contactless weight estimation of falling agglomerated particles. Our core contribution is to present and release the first specialized dataset for vision-based weight estimation of ground coffee: The \emph{Doppio} dataset comprises videos capturing a diverse range of roast types as they fall from a grinder at multiple grind settings, providing a benchmark to facilitate future research in this domain. Along with the dataset, we present a range of models, from a simple time-based model to recurrent deep neural networks, to showcase that weight information can be extracted from the continuous video signal.

\section{Related Work}

\subsubsection{Vision-Based Contactless Weight Estimation.}
To bypass the high cost and mechanical limitations of physical scales,
estimation of mass through computer vision finds application in broader agricultural and industrial contexts \cite{liang2024cvbased,mathiassen2011high,NYALALA2019288}. In livestock monitoring, various approaches for weight estimation have been introduced, demonstrating a shift toward contactless sensors to reduce operational complexity~\cite{dohmen2022computer}. In the context of vegetable and food processing, geometric feature mapping alongside multi-form shape properties has been used to estimate weights~\cite{kamiwaki2024machine,puri2009recognition}. 
Crucially, when dealing with materials in motion, recent work has explored tracking frameworks; for instance, deep learning-based object detection has been used to estimate the cumulative mass flow of irregular crops moving rapidly along a harvester conveyor belt~\cite{sunghyuk2023development}.
More broadly, deep learning-based end-to-end image-to-mass frameworks have successfully modeled the complex relationship between an item's visual geometry, volume, and hidden material densities to predict physical weight \cite{dehais2016two,standley2017image2mass} or calories \cite{he2013food,konstantakopoulos2024review,meyers2015im2calories,ruede2021multi}.

\subsubsection{Mass Flow Estimation of Particles.}
In the domain of mass flow estimation for particles in industrial settings, it is common to employ measurement methods that use modalities other than RGB imagery \cite{Samadi:2022:RSP}. Optical sensors, using photoreceptors and lasers, have been employed to estimate mass flow rates by measuring the length of falling particle clusters~\cite{Grift:2003:FMF}. For pharmaceutical tablet manufacturing, both capacitance-based sensing~\cite{Huang:2022:RMP} and X-ray sensors~\cite{Ganesh:2017:AXS} have demonstrated real-time, in-line mass flow rate monitoring in a non-invasive manner. Microwave Doppler radar has been applied to solid-flow measurements, with falling beans serving as a model~\cite{Isa:2006:MDR}. Using an image-based analysis, Gao \etal~\cite{Gao:2011:CIS} recover the particle size distribution from a laser-illuminated stream of pneumatically conveyed particles using a high-speed charge-coupled device camera and contour-based image processing. While these approaches demonstrate that reliable mass flow estimation and particle characterization are achievable in industrial pipelines, they generally rely on a pneumatically controlled mass flow or dedicated, specialized hardware (\eg, X-ray, solid-state laser, or microwave Doppler radar).

\subsubsection{Deep-Learning-Based Coffee Analysis.}
Translating visual volume into a precise weight metric requires accounting for particle size and material density. Due to limited access to industrial particle-flow systems, we utilize coffee grinding as an accessible case study. In the context of coffee processing, research has focused on analyzing the following granular variations. For example, coffee granularity classification has been approached using AlexNet~\cite{alexnet,Leonard:2022:CGS}.%
Exploring the use of simple consumer hardware \cite{Lertsawatwicha:2023:MPS} demonstrated that using traditional computer vision techniques, such as Canny edge detection \cite{canny} and connected components labeling \cite{Stefano:1999:CCL}, allows for measuring fine particles (\num{200}-\SI{1600}{\micro\meter}) using mobile-phone cameras. Following this approach, shape analysis was used on ground coffee to determine the distribution of particle sizes~\cite{Ren:2024:DIA}. 
Further work in the coffee industry focused on classifying raw materials \cite{DBLP:conf/icce-tw/ChenJH22} and on quality control \cite{hsia2022explainable,DBLP:conf/apwcs2/LiangXZYC23} during the roasting process. Several studies have leveraged deep learning architectures to categorize bean characteristics. For instance, a classification dataset comprising \SI{8}{k} images of unroasted coffee beans was proposed to identify defects and bean varieties~\cite{Febriana:2022:UCD}. Regarding the roasting process, standard deep learning-based vision architectures such as LeNet \cite{Lecun:1998:LEN}, ResNet \cite{resnet}, DenseNet \cite{Huang:2017:DNT}, and MobileNet\nobreakdash-V2 \cite{mobilenetv2} were trained and benchmarked on individual beans at various roast levels~\cite{Hassan:2024:ECB}. Similar classifications on images containing bulk quantities of roasted beans were also performed using neural networks~\cite{Leme:2019:RCR}. While these works establish the effectiveness of using deep learning-based vision models for identifying coffee-specific features, they remain limited to static, non-temporal analysis.

While these studies focus on analyzing the coffee, they do not quantify how much is being ground in a dynamic setting. In this work, we extend these concepts by moving beyond static, categorized, or volumetric analysis toward spatiotemporal understanding for efficient, real-time contactless weight estimation of falling particles.

\section{The \emph{Doppio} Dataset}
\begin{figure}
    \centering
    
\begin{tikzpicture}[
    spy using outlines={
        rectangle, 
        magnification=2, 
        size=1.98cm, 
        very thick, 
        every spy in node/.append style={fill=white, thick},
        every spy on node/.append style={very thick} 
    },
    sequence box/.style={draw=pipelineW, fill=white, thick, minimum width=1.5cm, minimum height=1cm, align=center, font=\sffamily},
    data frame/.style={draw=none, fill=none, thick, minimum width=2.85cm, minimum height=1.8cm, align=center, font=\sffamily},
    >={Stealth[inset=0pt,length=7.0pt,angle'=45]},
]
    \def\imgBack{00634.jpg}
    \def\imgMid{00632.jpg}
    \def\imgFront{00631.jpg}

    \node[inner sep=0pt, anchor=north west, draw=black, thick] at (-0.2cm, 10.2cm) {
        \includegraphics[height=5cm]{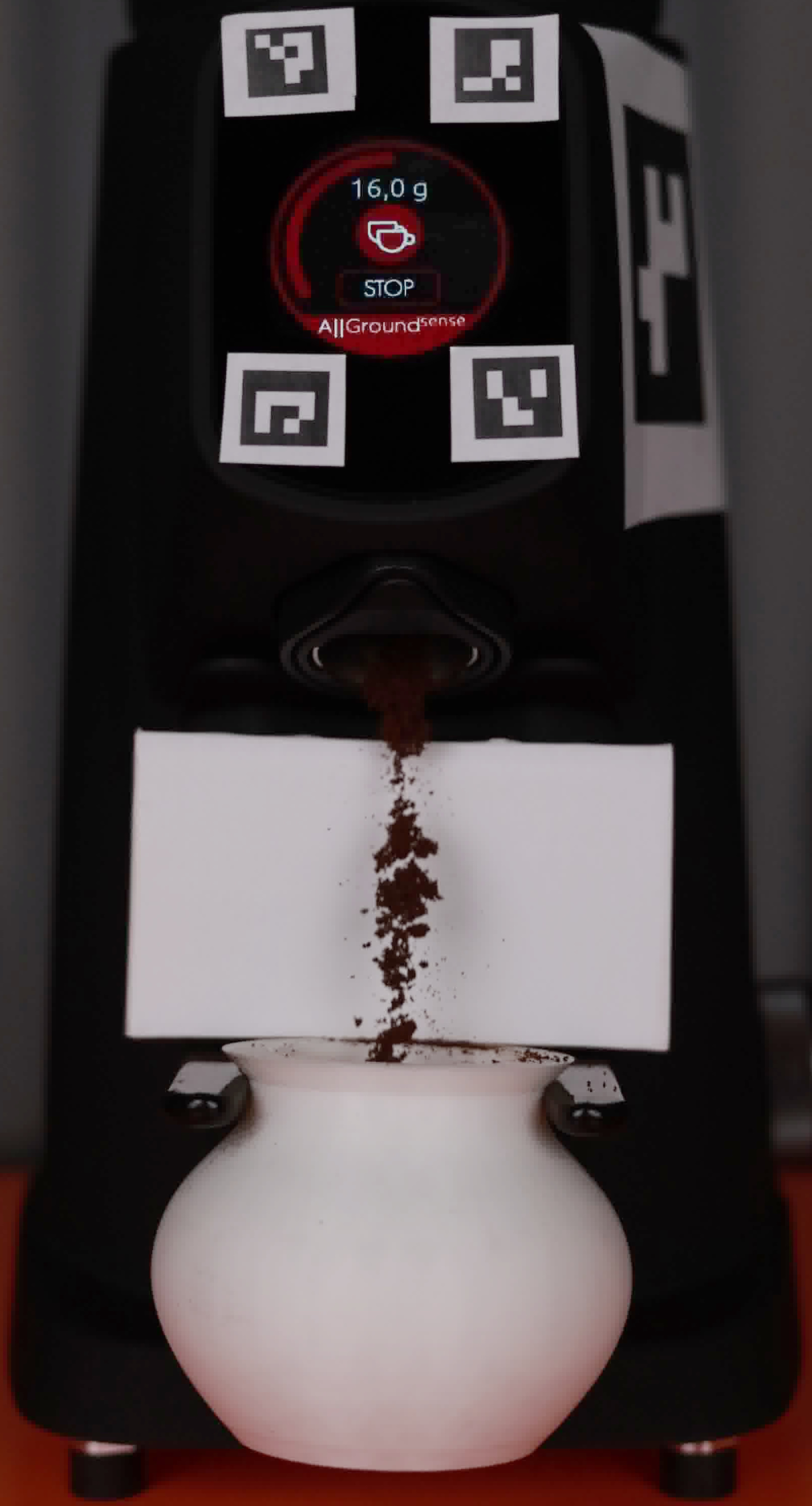}
    };
    \node[inner sep=0pt, anchor=north west, draw=black, thick] at (-0.1cm, 10.1cm) {
        \includegraphics[height=5cm]{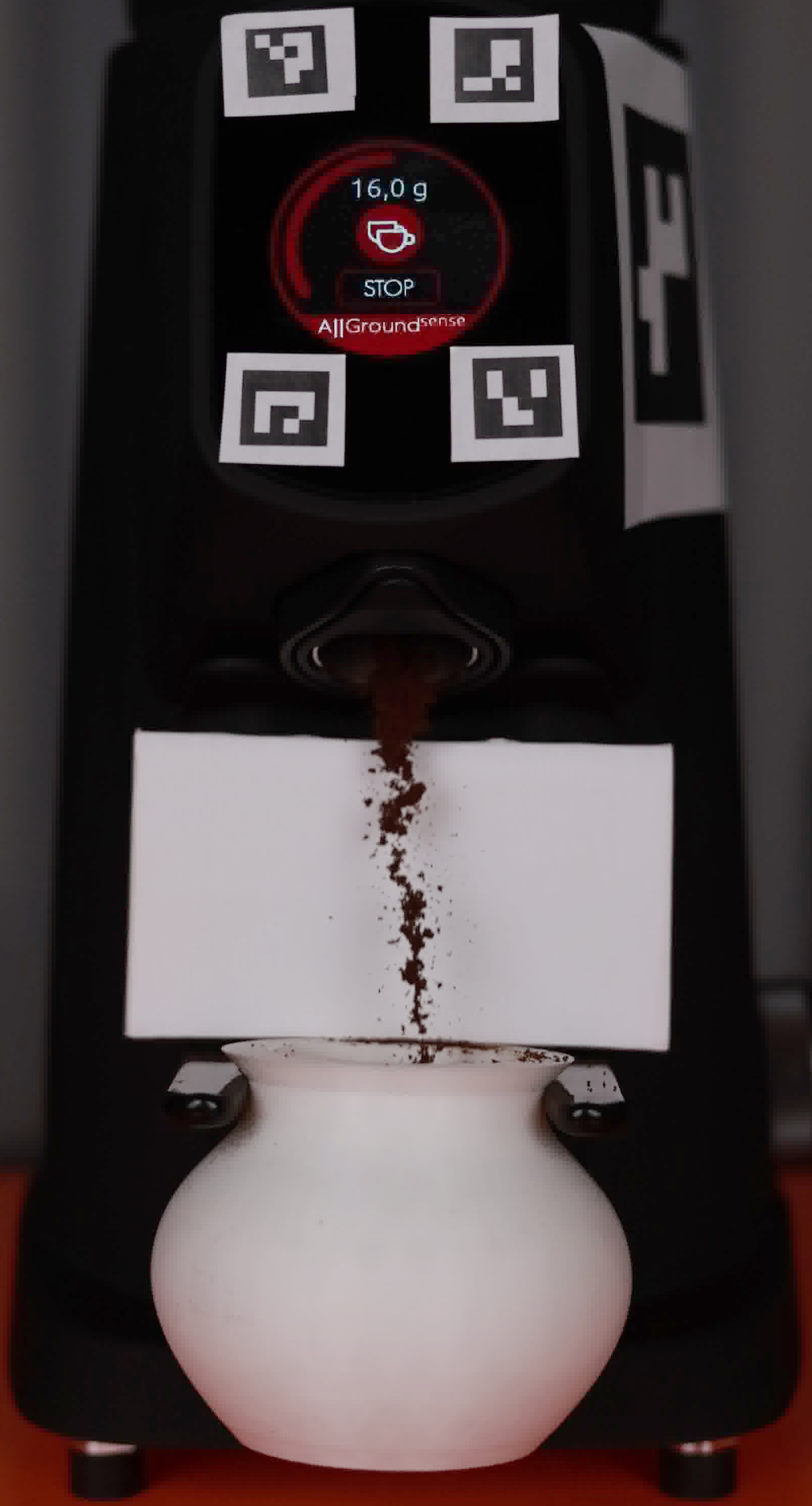}
    };
    \node (machine) [inner sep=0pt, anchor=north west, draw=black, thick] at (0cm, 10cm) {
        \includegraphics[height=5cm]{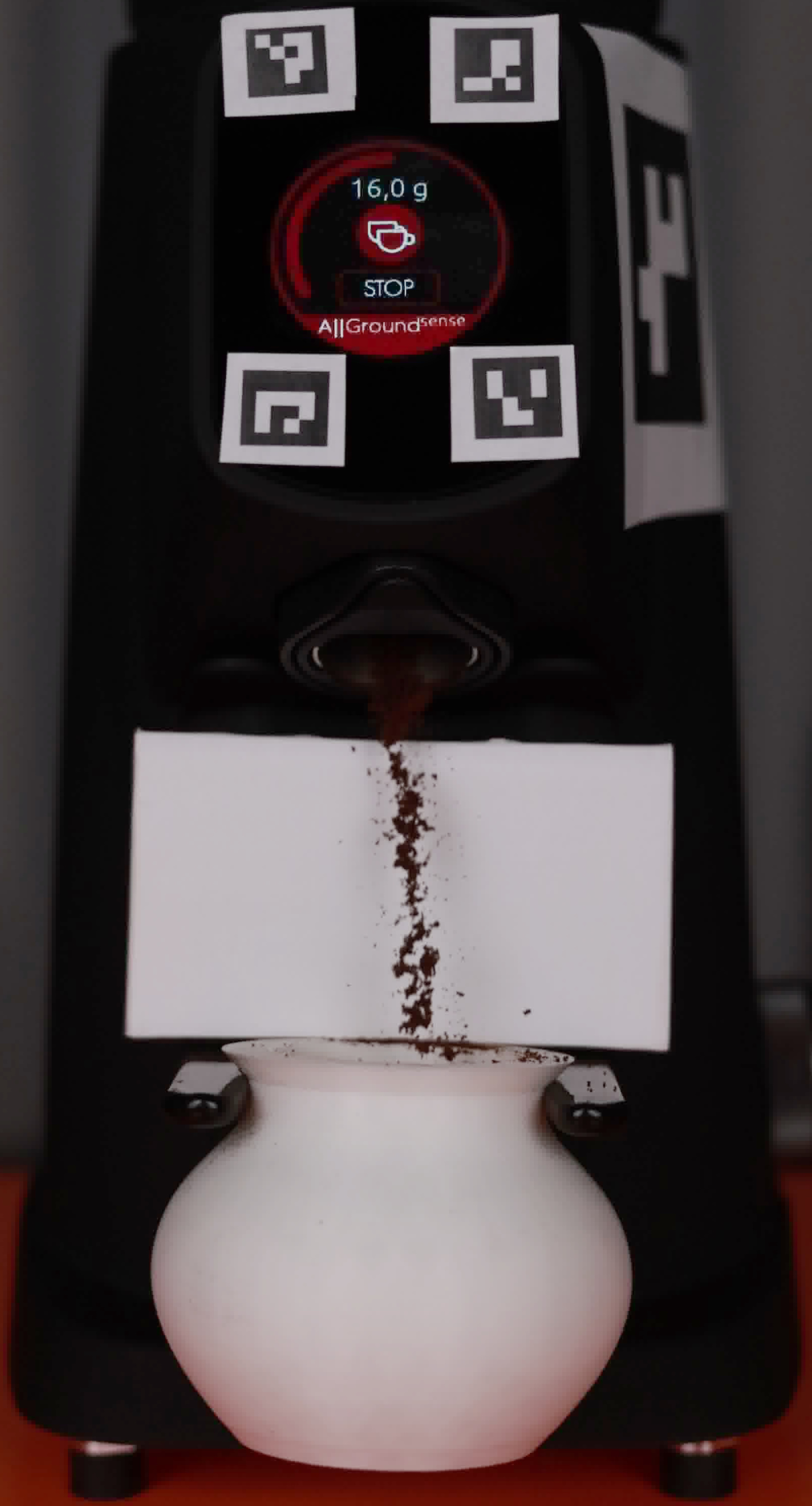}
    };

    \node[draw=pipelineW, thick, fill=white, minimum size=1.98cm,
          path picture={
              \node[anchor=north west, inner sep=0pt] at (-2.6cm, 1.70cm) {\includegraphics[height=10cm]{\imgBack}};
          }] at (4.1cm, 9.15cm) {};
          
    \node[draw=pipelineW, thick, fill=white, minimum size=1.98cm,
          path picture={
              \node[anchor=north west, inner sep=0pt] at (-2.6cm, 1.70cm) {\includegraphics[height=10cm]{\imgMid}};
          }] at (4.3cm, 8.95cm) {};
          
    \spy [pipelineW] on (1.3cm, 9.15cm) in node (displaySpy) at (4.5cm, 8.75cm);

    \node[draw=pipelineI, thick, fill=white, minimum size=1.98cm,
          path picture={
              \node[anchor=north west, inner sep=0pt] at (-2.8cm, 5.94cm) {\includegraphics[height=10cm]{\imgBack}};
          }] at (4.1cm, 6.40cm) {};
          
    \node[draw=pipelineI, thick, fill=white, minimum size=1.98cm,
          path picture={
              \node[anchor=north west, inner sep=0pt] at (-2.8cm, 5.94cm) {\includegraphics[height=10cm]{\imgMid}};
          }] at (4.3cm, 6.20cm) {};

    \spy [pipelineI] on (1.4cm, 7.03cm) in node (coffeeSpy) at (4.5cm, 6.00cm);

    \node[sequence box] (ocr3) at (7.15cm, 9.15cm) {};  
    \node[sequence box] (ocr2) at (7.35cm, 8.95cm) {};  
    \node[sequence box] (ocr1) at (7.55cm, 8.75cm) {16.0g};

    \draw[->, very thick, pipelineW] (5.49cm, 8.75cm) -- (ocr1.west);

    \draw[line width=1.7pt, line cap=round, dash pattern=on 0pt off 3pt, shorten >=0pt, shorten <=0pt] ([xshift=0.1cm, yshift=0.0cm]ocr3.north east) to node[midway, above, inner sep=1pt, font=\sffamily\scriptsize, sloped] {Smoothing} (ocr1.north east);

    \node[sequence box] (deltat) at (10.5cm, 8.75cm) {Time-Lag\\Compensation};

    \draw[->, very thick, pipelineW] (ocr1.east) -- (deltat.west);

    \node[data frame, opacity=0.2] (frame3) at (7.825cm, 6.60cm) {
        {\color{pipelineI}\Bigg(}%
        \tikz[baseline=-0.5ex]{
            \node[draw=none, thick, inner sep=0pt, minimum size=0.86cm, 
                  path picture={
                      \node at (0.0cm, 0.4cm) {\includegraphics[height=4.5cm]{\imgBack}};
                  }] {};
        }, 16.4g%
        {\color{pipelineW}\Bigg)}%
    };
    \node[data frame, opacity=0.3] (frame2) at (8.025cm, 6.30cm) {
        {\color{pipelineI}\Bigg(}%
        \tikz[baseline=-0.5ex]{
            \node[draw=none, thick, inner sep=0pt, minimum size=0.86cm, 
                  path picture={
                      \node at (0.0cm, 0.4cm) {\includegraphics[height=4.5cm]{\imgMid}};
                  }] {};
        }, 16.3g%
        {\color{pipelineW}\Bigg)}%
    };
    
    \node[data frame] (frame1) at (8.225cm, 6.00cm) {
        {\color{pipelineI}\Bigg(}%
        \tikz[baseline=-0.5ex]{
            \node[draw=none, thick, inner sep=0pt, minimum size=0.86cm, 
                  path picture={
                      \node at (0.0cm, 0.4cm) {\includegraphics[height=4.5cm]{\imgFront}};
                  }] {};
        }, 16.2g%
        {\color{pipelineW}\Bigg)}%
    };

    \node[anchor=north, font=\sffamily, inner sep=0pt] at (8.025cm, 5.37cm) {\emph{Doppio} Dataset};
    \node[anchor=north, font=\sffamily, inner sep=0pt] at (7.35cm, 8.1cm) {OCR Readings};

    \draw[->, very thick, pipelineI] (5.49cm, 6.00cm) -- ([xshift=-0.2cm]frame1.west);
    
    \draw[->, very thick, pipelineW] (deltat.south) |- (frame1.east);

\end{tikzpicture}

    \caption{\textbf{Data acquisition pipeline} for a single sequence of the \emph{Doppio} dataset. Crops of the display and coffee are taken relative to the ArUco markers. The display crop is then processed by an optical character recognition (OCR) approach, and a time-lag compensation is applied. Finally, we construct a sequence of image-weight pairs.} %
    \label{fig:dataset_capture}
\end{figure}

Due to the lack of publicly available datasets containing videos of falling particles with per-frame weight annotations, we record a dedicated dataset---\emph{Doppio}. We will release our \emph{Doppio} dataset, evaluation metrics, and data loader with the publication of this paper.
We use a coffee grinder to produce falling coffee particles and capture videos of the falling coffee. In these videos, we also capture the machine's display showing the cumulative weight of the coffee determined by an integrated scale.

\subsection{Data Acquisition}
For our recordings, we use a Fiorenzato AllGround Sense coffee grinder. This grinder offers discrete grind size adjustments and an integrated scale, well-suited for our data acquisition. The scale provides the real-time weight measurements, which we record to obtain ground-truth weight measurements. To ensure sufficient variability, we record sequences spanning \num{25} different grind sizes and \num{3} different types of coffee beans. Each video in the dataset shows a continuous grinding process that yields about \SI{32}{\gram} of ground coffee. \Cref{fig:dataset_capture} \emph{(left)} provides an overview of our data acquisition setup.

To ensure a consistent spatial arrangement of our capturing setup, we attach ArUco markers~\cite{Garrido-Jurado:2014:AGD} to all critical components in the scene. This includes the floor, the main camera, the grinder, the lighting, and the tripods. For robust pose estimation and tracking, each object is equipped with at least three markers. A Logitech C920 webcam is used to detect these markers and maintain spatial calibration between recordings.

To record the main sequence of falling particles, it is important to capture the fast-moving coffee particles without motion blur, so they remain clearly visible. Therefore, we employ a Canon EOS R5 camera, recording at \SI{60}{fps} at a \SI{4}{K} resolution, with a fast shutter speed of \qty[parse-numbers=false]{1/2000}{\second}, an aperture of $\text{f}/\text{\num{3.2}}$, and an ISO value of \num{1000}. In \cref{fig:dataset_capture}, we show an example frame captured by the camera with these settings.

\subsection{Postprocessing and Annotation}
We extract a \numproduct{512\,x\,512}\,\unit{\px} crop of the falling coffee from each frame. To determine the cumulative weight, we extract a second crop that contains the machine's display (\cf \cref{fig:dataset_capture} \emph{(middle)}). Then, we apply an optical character recognition (OCR) pipeline based on Tesseract OCR \cite{Smith:2007:OTO} to read the display's content. Since the weight is unreadable in some frames due to the display refreshing its content while the frame is captured, we filter out OCR misdetections and fill missing values. We identify misdetections by assuming a monotonic growth in weight and by limiting the maximum increase between consecutive frames. The missing values are then linearly interpolated from their closest valid neighbors.

We also compensate for the time lag %
between when the ground coffee appears in the camera frame and when the weight is measured (\cf \cref{fig:dataset_capture} \emph{(right)}). %
This offset is caused by the time it takes for the coffee to fall onto the scale after passing through the cropped window. To that end, we estimate and apply a per-sequence time offset. This offset is estimated per sequence using the first frame in which ground coffee appears and the frame in which the scale detects an initial weight.

\begin{figure}[t]
    \centering
    
\begin{tikzpicture}[spy using outlines={circle,magnification=5,size=2.4cm,connect spies,very thick, every spy in node/.append style={rotate=0, thick, fill=white}}]
     \tikzset{every picture/.style={/utils/exec={\sffamily\small}}}

    \begin{axis}[
        anchor=west,
        ymajorgrids,
        xtick pos=bottom,
        ytick pos=left,
        cycle list/Pastel1,
        width=1.02\figureWidth,
        height=45mm,
        xlabel={Frame},
        ylabel={Weight / g},
        xlabel shift=-3pt,
        xmin=1,
        xmax=1328,
        ymin=0,
        ymax=33.0,
        legend style={
            font=\small, 
            cells={anchor=west}, 
            at={(0.5,1)},
            legend columns=2,
            anchor=center
        },
        legend image post style={line width=2pt},
        xtick={200, 400, 600, 800, 1000, 1200},
        xticklabels={200, 400, 600, 800, 1000, 1200},
        ytick={0, 10, 20, 30},
        yticklabels={0, 10, 20, 30},
        ticklabel style = {font=\small},
    ]
        \fill [Paired-A, opacity=0.2]
            (1, \pgfkeysvalueof{/pgfplots/ymin}) rectangle (256, \pgfkeysvalueof{/pgfplots/ymax});
        \node [below, yshift=-0.65cm, font=\small, text=Pastel2-C!50!black] 
            at (128, \pgfkeysvalueof{/pgfplots/ymax}) {start};

        \fill [Paired-G, opacity=0.2]
            (436, \pgfkeysvalueof{/pgfplots/ymin}) rectangle (692, \pgfkeysvalueof{/pgfplots/ymax});
        \node [below, yshift=-0.65cm, font=\small, text=Pastel2-B!50!black] 
            at (564, \pgfkeysvalueof{/pgfplots/ymax}) {mid};

        \fill [Paired-C, opacity=0.2]
            (1072, \pgfkeysvalueof{/pgfplots/ymin}) rectangle (1328, \pgfkeysvalueof{/pgfplots/ymax});
        \node [below, yshift=-0.65cm, font=\small, text=Pastel2-A!50!black] 
            at (1200, \pgfkeysvalueof{/pgfplots/ymax}) {end};
        
        \addplot+[line width=0.8pt, Paired-E!70!white] table [
            col sep=semicolon,
            x=timestep,
            y=raw
        ] {dataset_weights.csv};
        \addlegendentry{Raw Weight}

        \addplot+[line width=0.3pt, Paired-J] table [
            col sep=semicolon,
            x=timestep,
            y=smooth
        ] {dataset_weights.csv};
        \addlegendentry{Smoothed Weight}

        \coordinate (spypoint) at (590, 14.8); 
        \coordinate (magnifyglass) at (rel axis cs:0.68, 0.46);

    \end{axis}

    \spy[black] on (spypoint) in node at (magnifyglass);
\end{tikzpicture}

    \vspace{-0.75em}
    \caption{\textbf{Example weight sequence.} Visualization of the raw and smoothed weights of one exemplary sequence of our \emph{Doppio} dataset. Additionally, the \textit{start}\;\colorindicator{Paired-A}, \textit{mid}\;\colorindicator{Paired-G}, and \textit{end}\;\colorindicator{Paired-C} sections are highlighted.}
    \label{fig:weight_smoothing}
\end{figure}
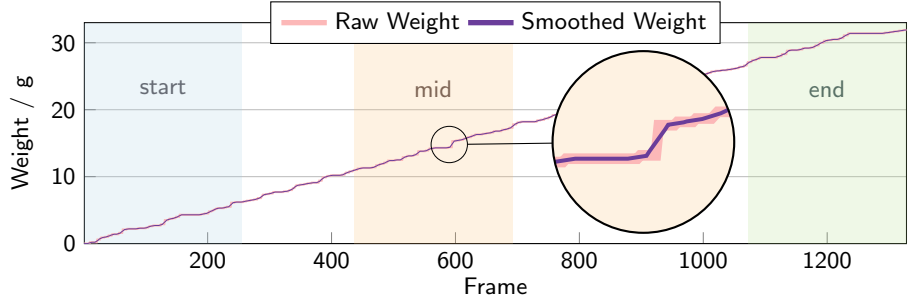

Since the integrated scale measurements show only a single digit after the decimal point, we apply a moving average window of seven frames to the raw measurements as a smoothing operator. This avoids ramp-like trajectories in our data. An illustrative example of this smoothing effect is presented in \cref{fig:weight_smoothing}.%

\subsection{Dataset Splits}
We split our dataset into \num{131} training, \num{13} validation, and \num{75} test sequences. Further statistics of the splits are reported in \cref{tab:dataset_stats}. We ensure each combination of grind size and coffee bean type is present in the training and test set.

\begin{table}[t]
    \centering
    \caption{\textbf{Dataset statistics overview.} We report statistics for training, validation, and test splits: mean weight increment and standard deviation per frame, minimum and maximum number of frames per sequence, and number of sequences per split.}
    
\footnotesize%
\renewcommand{\arraystretch}{0.975}
\hspace{-3.3pt}\begin{tabularx}{\linewidth}{X*{2}{@{\hspace{4.5pt}}S[table-format=1.4,round-mode=places,round-precision=4]}@{\hspace{4.5pt}}S[table-format=3.0]@{\hspace{4.5pt}}S[table-format=4.0]@{\hspace{4.5pt}}S[table-format=3.0]}
        \toprule
        \textbf{Split} & {\textbf{Mean (g/frame)}} & {\textbf{Std (g/frame)}} & {\textbf{Min frames}} & {\textbf{Max frames}} & {\textbf{Sequences}} \\
        \midrule
        train     & 0.03041 & 0.01930 & 810  & 1406 & 131 \\
        val  & 0.02838 & 0.01960 & 979  & 1416 &  13 \\
        test & 0.03029 & 0.01924 & 839  & 1480 &  75 \\
        \bottomrule
    \end{tabularx}

    \label{tab:dataset_stats}
\end{table}

We further split the validation and test sequences into the following sub-sequences: \textit{Start} and \textit{end} contain the first and last \num{256} frames of each sequence, respectively. \textit{Mid} contains \num{256} consecutive frames sampled from in-between the start and end sections (\cf \cref{fig:weight_smoothing}). The \textit{full} setting contains the entire sequence. These sub-sequences enable more detailed model evaluations, as the grinder's behavior varies throughout the grinding process. For example, at the start, slightly fewer grounds fall than in the middle. Towards the end, the machine's internal control loop stops or slows down the grinding process to reach the predefined target weight more accurately (\cf \cref{fig:weight_smoothing} \emph{(right top)}). We include a video of the complete grinding process in our supplemental material to demonstrate this.

\subsection{Evaluation Metrics}
For our evaluation metrics, we define a doppio (a double shot of espresso, a common espresso serving) as \SI{16}{\gram} of ground coffee. We measure the mean absolute error (MAE) per doppio. We term this metric \MAED, and we define it as
\begin{equation}
    \text{\MAED} = \frac{16}{N} \sum_{i=1}^N \frac{|w_i(L_i) - \hat{w_i}(L_i)|}{w_i(L_i)},
\end{equation}
where $N$ is the number of sample sequences, $w_i(j)$ is the measured ground truth weight at the $j$-th frame for the $i$-th sequence. $\hat{w_i}(j)$ denotes the estimated weight. $L_i$ is the length of the $i$-th sequence, and therefore $w_{i}(L_i)$ refers to the last element of the sequence, so the MAE is only calculated after the entire sequence is processed.

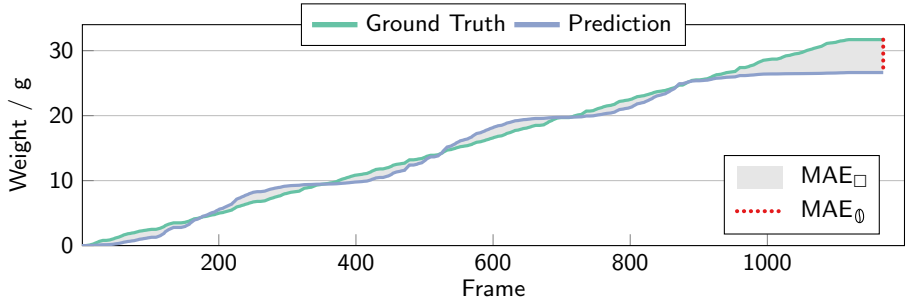
\begin{figure}[t]
    \centering
    
 \tikzset{every picture/.style={/utils/exec={\sffamily\small}}}
\begin{tikzpicture}[spy using outlines={circle,magnification=5,size=2.4cm,connect spies,very thick, every spy in node/.append style={rotate=0, thick, fill=white}}]

    \begin{axis}[
        anchor=west,
        ymajorgrids,
        xtick pos=bottom,
        ytick pos=left,
        cycle list/Pastel1,
        width=1.02\figureWidth,
        height=45mm,
        xlabel={Frame},
        ylabel={Weight / g},
        xlabel shift=-3pt,
        xmin=1,
        xmax=1200,
        ymin=0,
        ymax=34,
        xtick={200, 400, 600, 800, 1000},
        xticklabels={200, 400, 600, 800, 1000},
        ytick={0, 10, 20, 30},
        yticklabels={0, 10, 20, 30},
        ticklabel style = {font=\small},
        legend style={
            font=\small, 
            cells={anchor=west}, 
            at={(0.5,1)},
            legend columns=2,
            anchor=center
        },
        legend image post style={line width=2pt},
    ]
        \addplot+[line width=1.2pt, Set2-A, name path=gt_curve] table [
            col sep=semicolon,
            x=timestep,
            y=gt
        ] {metric_sample.csv}coordinate [pos=1] (gt_end);
        \addlegendentry{Ground Truth}

        \addplot+[line width=1.2pt, Set2-C, name path=pred_curve] table [
            col sep=semicolon,
            x=timestep,
            y=pred
        ] {metric_sample.csv} coordinate [pos=1] (pred_end);
        \addlegendentry{Prediction}

        \addplot [
            fill=gray!20,
            draw=none,
            forget plot
        ] fill between [
            of=gt_curve and pred_curve
        ];

        \draw [Set1-A, line width=1.5pt, line cap=round, dash pattern=on 0pt off 2.5pt] (gt_end) -- (pred_end);
        
        \node [
            draw=black,          
            fill=white,          
            anchor=south east,   
            font=\scriptsize,    
            inner sep=4pt        
        ] at (rel axis cs:0.97, 0.05) { 
            \begin{tabular}{@{}c@{\hspace{6pt}}l@{}}
                
                \tikz[baseline=-0.7ex]{
                    \fill[gray!20, draw=none]
                    (0, -0.15cm) rectangle (0.6cm, 0.15cm);
                } & \small \MAEA \\[4pt]
                
                \tikz[baseline=-0.7ex]{
                    \draw[Set1-A, line width=1.5pt, line cap=round, dash pattern=on 0pt off 2.5pt] (0, 0) -- (0.6cm, 0);
                } & \small \MAED \\
                
            \end{tabular}
        };
    \end{axis}
\end{tikzpicture}

    \vspace{-0.75em}
    \caption{\textbf{Metrics visualization.} \MAEA (gray shaded area\;\colorindicator{gray!20}) is calculated between the prediction and ground truth for each frame. \MAED (red\;\colorindicator{Set1-A} dashed line) is computed at the end of the sequence and measures the final discrepancy between the predicted and ground truth weight, while normalizing to the discrepancy per doppio (\ie, \SI{16}{\gram}).}
    \label{fig:metrics}
\end{figure}

Since our dataset contains per-frame weight annotations, we can measure the MAE at each time step and calculate the area between the predicted and ground-truth weight measurements. To enable comparison of variable-length sequences, we normalize this metric by $N$, the number of frames within a sequence. We define this metric as
\begin{equation}
    \text{\MAEA} = \frac{1}{N} \sum_{i=1}^N \frac{1}{L_i} \sum_{j=1}^{L_i} |w_i(j) - \hat{w_i}(j)|.
\end{equation}
We provide a visualization of %
how %
\MAED and \MAEA relate to the sequence of frames in \cref{fig:metrics}.

\section{Methods for Contactless Weight Estimation}
To systematically address the challenge of contactless weight estimation, we evaluate models of various complexity, ranging from simple linear regression to deep spatio-temporal architectures. First, we establish a naïve, time-based baseline to quantify the predictive boundaries achievable without any visual inputs, followed by more sophisticated computer vision models. %

\subsection{Time-Based Baseline}
The naïve approach to estimating the weight of falling particles is a time-based model calibrated for each new combination of input material and particle size. We build this baseline model using linear regression for each possible combination of bean type and grind size, enabling us to assess whether more complex vision-based models are needed or whether the assumption of a constant fall rate is sufficient to solve this task.

\subsection{Vision-Based Models}
\label{sec:vision_models}
To predict the target weights across our video data, we process each frame individually to estimate the weight difference caused by the coffee visible at that time step. Per-frame differences are then accumulated using a cumulative sum to obtain the total weight of the processed sequence up to any given frame. For our task of estimating the cumulative weights of falling particles, we explore three vision-based approaches: a feed-forward approach that accumulates no temporal information, a recurrent architecture that can capture long sequences, and a temporal approach that considers multiple previous frames for its predictions. Since the weight can only increase monotonically over time, we limit our models to output only positive weight estimates by using a softplus~\cite{Dugas:2000:ISF} activation function.

As inputs for all of our vision-based models, we use either a single RGB frame $I_t\in \mathbb{R}^{H\times W\times C}$ with $C = \text{\num{3}}$ or a concatenation of the current frame and the pixel-wise difference between the current and previous frame, denoted as $\Delta I_t$, doubling the channel dimension to $C = \text{\num{6}}$.

Our feed-forward approach uses a standard feed-forward network (FFN) to directly predict the weight difference from only the input at the current time step, ignoring all previous frames.
Standard FFNs lack temporal awareness, meaning that falling particles visible in multiple frames can skew the results. Thus, we also employ a recurrent architecture to track these particles accurately over time.
We aim to capture long-term temporal dependencies by employing a gated recurrent unit (GRU)~\cite{Cho:2014:GRU} coupled with a feed-forward backbone acting as a feature extractor. The extracted spatial features are passed to the GRU, which updates its hidden state and outputs the predicted weight differences.
Since long-term temporal context may be unnecessary for this task, we also evaluate temporal convolutional networks (TCNs) \cite{Lea:2017:TCN} as an alternative to GRUs, using the same spatial features.
In the TCNs, we apply causal convolutions along the time axis to also consider information from previous frames within a limited time window.

\section{Experiments}
\label{sec:experiments}
\subsection{Training Setup}

For all experiments, we train our deep-learning-based models on sub-sampled sequences of \num{128} consecutive frames within each batch. Our training runs for \num{200} epochs using a one-cycle learning rate schedule with cosine annealing~\cite{Smith:2019:SCV}, with the first \SI{10}{\%} of iterations used for warmup.

To improve robustness and generalization, we apply random adjustments to the brightness and contrast of each sequence. Furthermore, we introduce a domain-specific temporal augmentation by randomly inserting frames containing no falling coffee into \SI{10}{\%} of the frames of each train sequence. This encourages the network to correctly predict a zero-weight delta when no particles are present.

As the learning objective, we employ a combined loss that supervises both the final accumulated weight and the intermediate per-frame differences, enforcing correct temporal dynamics. Our loss on the \emph{last} accumulated weight is
\begin{equation}
    \mathcal{L}_{\text{seq}} = d\!\left(w_i(L_i), \hat{w_i}(L_i)\right),
    \label{eq:lossseq}
\end{equation}
where $d(x, y)$ measures the distance between $x$ and $y$. $w_i(L_i)$ is the ground truth weight and $\hat{w_i}(L_i)$ the estimated weight, both at the end of the $i$-th sequence denoted by $L_i$. To enforce correct temporal dynamics, we compute a difference loss between the predicted weight deltas and the ground-truth \emph{per frame} $j$:
\begin{equation}
    \mathcal{L}_{\delta} = \frac{1}{L_i - 1} \sum_{j=2}^{L_i} d\!\left(w_i(j) - w_i(j-1), \hat{w_i}(j) - \hat{w_i}(j-1)\right).
    \label{eq:lossdelta}
\end{equation}
Our final loss is a combination of $\mathcal{L}_{\text{seq}}$ and $\mathcal{L}_{\delta}$, weighted by a combination factor $\lambda\in\mathbb{R}^{+}$, \ie, $\mathcal{L} = \mathcal{L}_{\text{seq}} + \lambda \mathcal{L}_{\delta}$.
As the expected weights in our setting are numerically small, we use the smooth L1~\cite{Girshick:2015:FRC} distance for $d(x, y)$ across all experiments. An evaluation of alternative distance functions is provided in the supplement. 

Since all of our models described in \cref{sec:vision_models} use a feed-forward network for feature extraction, we finetune ImageNet-pretrained~\cite{imagenet} ResNets~\cite{resnet} and MobileNet-V3s~\cite{Howard:2019:mobilenetv3} of different sizes on our proposed dataset. For our GRU architectures, we use a single GRU layer with a hidden dimension size of \num{256}. In our TCNs, we use \num{4} causal convolution blocks with kernel size \num{5}, hidden dimension \num{256}, and a dilation that doubles in each block, starting at \num{1}.

\subsection{\emph{Doppio} Evaluation Results}
\begin{table}[t]
    \centering
    \caption{\textbf{Results on different input representations.} We report results of our models with and without frame difference features $\Delta I_t$ on \emph{Doppio} test, using \MAED and \MAEA (both $\downarrow$) for the \textit{full} sequences. We highlight the best results for each metric \emph{per row} in orange \colorindicator{best}.}
    
\footnotesize%
\setlength{\tabcolsep}{3.0pt}%
\renewcommand{\arraystretch}{0.925}
\hspace{-3.3pt}\begin{tabularx}{\figureWidth}{c@{\hspace{5pt}} | @{\hspace{4pt}} X S[table-format=1.2,round-mode=places,round-precision=2] S[table-format=1.2,round-mode=places,round-precision=2] l S[table-format=1.2,round-mode=places,round-precision=2] S[table-format=1.2,round-mode=places,round-precision=2]}
        \toprule
        \multicolumn{1}{c}{} & \multirow{2}{*}[-0.5ex]{\textbf{Backbone}} & \multicolumn{2}{c}{\textbf{Without} $\Delta I_t$\;\;} & \hphantom{04} &\multicolumn{2}{c}{\textbf{With} $\Delta I_t$}\\
        \cmidrule(lr{14pt}){3-4} \cmidrule(r){6-7}
        \multicolumn{1}{c}{} & & {\textbf{\MAED}} & {\textbf{\MAEA}} & & {\textbf{\MAED}} & {\textbf{\MAEA}}\\
        \midrule
        
        \multirow{4}{*}{\rotatebox[origin=c]{90}{\textbf{FFN}}}
        & MobileNet-V3-S & 0.28153669834136963 & 0.3908247649669647 & & \bestcell0.19478078186511993 & \bestcell0.25227487087249756\\
        & MobileNet-V3-L & 0.29154008626937866 & 0.37098821997642517 & &\bestcell0.19811062514781952 & \bestcell0.26023930311203003\\
        & ResNet-18 & \bestcell0.2430 & 0.3255 & & 0.2611 & \bestcell0.3003 \\
        & ResNet-34 & 0.2482869029045105 & 0.33045533299446106 & & \bestcell0.2095007449388504 & \bestcell0.24105773866176605\\
        \midrule
        
        \multirow{4}{*}{\rotatebox[origin=c]{90}{\textbf{GRU}}}
        & MobileNet-V3-S & \bestcell0.3457910120487213 & 0.32488223910331726 & & \bestcell0.34566011905670166 & \bestcell0.3103947341442108\\
        & MobileNet-V3-L & 0.24048636853694916 & 0.2587810754776001 & & \bestcell0.20696210861206055 & \bestcell0.22718612849712372\\
        & ResNet-18 & \bestcell0.1929 & 0.2616 & &  0.1967 & \bestcell0.2411\\
        & ResNet-34 & \bestcell0.19200077652931213 & \bestcell0.25547683238983154 & & 0.2682589292526245 & 0.32739055156707764\\
        \midrule

        \multirow{4}{*}{\rotatebox[origin=c]{90}{\textbf{TCN}}}
        & MobileNet-V3-S & 0.3105 & 0.3723 & & \bestcell0.2403 & \bestcell0.3096\\
        & MobileNet-V3-L & \bestcell0.1910 & \bestcell0.2274 & & 0.2852 & 0.3260\\
        & ResNet-18 & \bestcell0.1798 & \bestcell0.2321 & & \bestcell0.1752 & 0.2406\\
        & ResNet-34 & 0.1786 & \bestcell0.2325 & & \bestcell0.1670 & \bestcell0.2274\\
        \bottomrule
    \end{tabularx}

    \label{tab:ablations_delta}
\end{table}

\subsubsection{Design Choices.}
\label{sec:design}
As described in \cref{sec:vision_models}, we propose two input modes for our approaches: one that uses only the current frame $I_t$, and another that uses the current frame and its pixel-wise difference with the previous frame, denoted as $\Delta I_t$, stacked along the channel dimension. In \cref{tab:ablations_delta}, we analyze the impact of this design decision using different architectures in combination with various-sized backbones. For our feed-forward architectures, we find that using frame differences consistently improves the accuracy of all tested backbones except for ResNet-18 at very low computational cost, as only the first layer of each backbone needs to be modified to accommodate the six input channels. Given these results, we use the frame differences as additional inputs for all subsequent FFNs.
The differences between the results of our GRU-based architectures in \cref{tab:ablations_delta} are less pronounced, but given the large gap in accuracy for ResNets and the rather small gap for MobileNets, we chose to use the simpler inputs without frame differences in our subsequent GRU-based architectures. Analogously, for the TCN-based architectures, we observe a significant improvement in the accuracy of the MobileNet-V3-L backbone when the frame differences $\Delta I_t$ are excluded. Therefore, we decided not to use them in this setting.

We further investigate the choice of loss weighting factor $\lambda$ for the different architecture-types and find that $\lambda=\text{\num{10}}$ performs best for the feed-forward network, while $\lambda=\text{\num{100}}$ is best for GRU- and TCN-based architectures. Therefore, we use these differing hyperparameters for all subsequent experiments. For completeness, we provide the full tables for these experiments in the supplement. %

\begin{table}[t]
    \centering
    \caption{\textbf{Results on \emph{Doppio}.} We report results of our models on \emph{Doppio} test, using \MAED and \MAEA (both $\downarrow$) for different sections and for the \emph{full} sequences. Best results are highlighted for our FFNs, GRUs, and TCNs individually in orange \colorindicator{best} and the overall best results over all of our models in red \colorindicator{overallbest}.}
    
\setlength{\tabcolsep}{1.0pt}%
\renewcommand{\arraystretch}{0.925}
\footnotesize%
\hspace{-3.3pt}\begin{tabularx}{\linewidth}{c@{\hspace{5pt}} | @{\hspace{4pt}} X *{4}{S[table-format=1.2,round-mode=places,round-precision=2] S[table-format=1.2,round-mode=places,round-precision=2]}}
        \toprule
        \multicolumn{1}{c}{} & \multirow{2}{*}[-0.5ex]{\textbf{Method}} & \multicolumn{2}{c}{\textbf{Full}} & \multicolumn{2}{c}{\textbf{Start}}  & \multicolumn{2}{c}{\textbf{Mid}} & \multicolumn{2}{c}{\textbf{End}} \\
        \cmidrule(lr){3-4} \cmidrule(lr){5-6} \cmidrule(lr){7-8} \cmidrule(lr){9-10}
        \multicolumn{1}{c}{} & & {\textbf{MAE\textsubscript{\coffeebsmall}}} & {\textbf{MAE\textsubscript{$\square$}}} & {\textbf{MAE\textsubscript{\coffeebsmall}}} & {\textbf{MAE\textsubscript{$\square$}}} & {\textbf{MAE\textsubscript{\coffeebsmall}}} & {\textbf{MAE\textsubscript{$\square$}}} & {\textbf{MAE\textsubscript{\coffeebsmall}}} & {\textbf{MAE\textsubscript{$\square$}}} \\
        \midrule
        
        \multicolumn{1}{c}{} & Time baseline & 0.5352 & 0.5664 & 0.9828 & 0.2497 & 0.7465 & 0.2006 & 2.9520 & 0.6798 \\
        \midrule
        
        \multirow{4}{*}{\rotatebox[origin=c]{90}{\textbf{FFN}}} 
        & MobileNet-V3-S & \bestcell0.1948 & 0.2523 & \bestcell0.2916 & \overallbestcell0.1193 & 0.2850 & 0.1323 & \bestcell0.5692 & 0.1161 \\
        & MobileNet-V3-L & 0.1981 & 0.2602  & \bestcell0.2941 & \overallbestcell0.1240 & 0.2823 & 0.1313 & \bestcell0.5711 & \overallbestcell0.1142 \\
        & ResNet-18 & 0.2611 & 0.3003 & 0.3598 & 0.1380 & 0.2834 & 0.1321 & 0.6385 & 0.1274\\
        & ResNet-34 & 0.2095 & \bestcell0.2411 & 0.3234 & 0.1289 & \bestcell0.2349 & \overallbestcell0.1244 & 0.6261 & 0.1279 \\
        \midrule
        
        \multirow{4}{*}{\rotatebox[origin=c]{90}{\textbf{GRU}}} 
        & MobileNet-V3-S & 0.3458 & 0.3249 & 0.3701 & 0.1346 & 0.3178 & \bestcell0.1345 & 0.7534 & 0.1324 \\
        & MobileNet-V3-L & 0.2405 & \bestcell0.2588 & \bestcell0.3071 & \overallbestcell0.1234 & 0.2795 & \bestcell0.1318 & 0.5882 & \bestcell0.1210 \\
        & ResNet-18 & \bestcell0.1929 & \bestcell0.2616 & 0.3468 & 0.1304 & \bestcell0.2405 & \bestcell0.1287 & \bestcell0.5521 & \bestcell0.1156 \\
        & ResNet-34 & \bestcell0.1920 & \bestcell0.2555 & \bestcell0.3052 & \overallbestcell0.1239 & 0.2585 & \bestcell0.1308 & 0.5591 & \bestcell0.1191 \\
        \midrule
        
        \multirow{4}{*}{\rotatebox[origin=c]{90}{\textbf{TCN}}} 
        & MobileNet-V3-S & 0.3105 & 0.3723 & 0.4400 & 0.1553 & 0.3539 & 0.1427 & 0.5111 & \overallbestcell0.1142 \\
        & MobileNet-V3-L & 0.1910 & \overallbestcell0.2274 & \overallbestcell0.2730 & \overallbestcell0.1191 & 0.2358 & \overallbestcell0.1200 & 0.4747 & \overallbestcell0.1055 \\
        & ResNet-18 & \overallbestcell0.1798 & \overallbestcell0.2321 & 0.3063 & \overallbestcell0.1200 & 0.2323 & \overallbestcell0.1209 & 0.4705 & \overallbestcell0.1085\\
        & ResNet-34 & \overallbestcell0.1786 & \overallbestcell0.2325 & 0.2975 & \overallbestcell0.1216 & \overallbestcell0.2239 & \overallbestcell0.1177 & \overallbestcell0.4582 & \overallbestcell0.1066\\
        \bottomrule
    \end{tabularx}

    \label{tab:main_results}
\end{table}%

\subsubsection{Main Results.}
\cref{tab:main_results} presents our main performance evaluation. All of our proposed vision-based architectures consistently outperform the \emph{Time baseline} across all sub-sequences and metrics. When analyzing the purely spatial feed-forward networks (FFNs), the MobileNet-V3-S architecture achieves the highest overall accuracy. However, the performance margins are small, as the MobileNet-V3-L and ResNet-34 architectures achieve nearly identical accuracies.

When moving to temporal models, the trend shifts towards larger backbones, as all ResNet-based models outperform their MobileNet counterparts. In the case of GRUs, all MobileNet-based models are worse than in the FFN setting. This is likely due to the aggressive feature reduction, which, on the one hand, allows for high compute efficiency in MobileNets, but, on the other hand, limits the models' outputs to rather coarse features, lacking fine-grained details. For ResNets, we observe higher overall accuracies when used as backbones for GRUs and TCNs than when used as backbones for FFNs. In fact, our overall best model is a TCN using a ResNet-34 backbone, scoring a \MAED of \num{0.18} on the full test sequence.

To confirm that our models track only the visible coffee per frame, we analyze our TCN (w/ ResNet-34) using Grad-CAM~\cite{Selvaraju:2020:VED}.
\Cref{fig:gradcam} shows qualitative Grad-CAM results. In general, the model only pays attention to falling coffee if present in the frame, and it only relies on cues from the background when no coffee is present. However, the last sample also shows that the network sometimes only pays attention to larger lumps as they fall, ignoring smaller ones.

\begin{table}[t]
    \centering
    \caption{\textbf{Leave-one-out validation.} We train our TCN (w/ ResNet-34) only on training samples of certain types of coffee beans and report the overall \MAED ($\downarrow$) on the \emph{full} test sequences of \emph{Doppio}. We also evaluate each type of bean individually and report the difference to the overall \MAED indicated in gray \colorindicator{darkgray!75}.}
    
\footnotesize%
\setlength{\tabcolsep}{0.75pt}%
\renewcommand{\arraystretch}{0.925}
\hspace{-3.3pt}\begin{tabularx}{\figureWidth}{
        X S[table-format=1.2,round-mode=places,round-precision=2]c
        S[table-format=1.2,round-mode=places,round-precision=2]lc
        S[table-format=1.2,round-mode=places,round-precision=2]lc
        S[table-format=1.2,round-mode=places,round-precision=2]l
    }
        \toprule
        \textbf{Training Split} & {\textbf{All}} & \hphantom{0701} & \multicolumn{2}{c}{\textbf{A}} & \hphantom{0701} & \multicolumn{2}{c}{\textbf{B}} & \hphantom{0701} & \multicolumn{2}{c}{\textbf{C}} \\
        \midrule
        \{A, B, C\}  & 0.17859339714050293 & & 0.17598561064610116 & \diff{+\num{0.00}} & & 0.17811185495367102 & \diff{+\num{0.00}} & & 0.1806980847461485 & \diff{+0.00}\\
        \midrule
        \{A, B\}     & 0.22207146726 & & 0.2271299989936525 & \diff{+\num{0.01}} & & 0.19807069400713773 & \diff{-\num{0.02}} & & 0.2410137087818512 & \diff{+\num{0.02}}\\
        \{A, C\}     & 0.38492295251 & & 0.3092522897221073 & \diff{-\num{0.07}} & & 0.43523807550138666 & \diff{+\num{0.06}} & & 0.41027849231261193 & \diff{+\num{0.03}}\\
        \{B, C\}     & 0.28194567278 & & 0.3049199406185363 & \diff{+\num{0.02}} & & 0.265719623349373 & \diff{-\num{0.01}} & & 0.27519745437261295 & \diff{+\num{0.00}}\\
        \bottomrule
    \end{tabularx}

    \label{tab:loocv}
\end{table}%
\begin{figure}[t]
    \centering

\setlength{\tabcolsep}{1.365pt}%
\begin{tabularx}{\linewidth}{@{}cccccc@{}}
        \includegraphics[width=0.1615\textwidth]{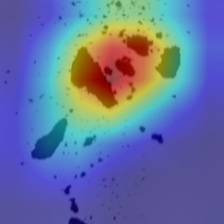} & 
        \includegraphics[width=0.1615\textwidth]{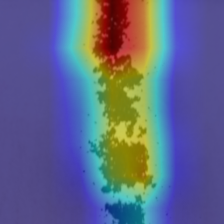} & 
        \includegraphics[width=0.1615\textwidth]{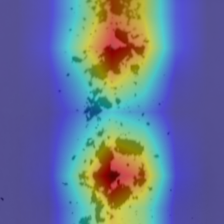} & 
        \includegraphics[width=0.1615\textwidth]{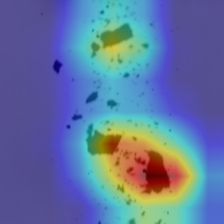} &
        \includegraphics[width=0.1615\textwidth]{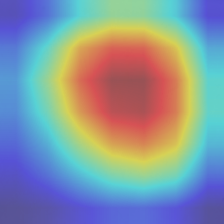} &
        \includegraphics[width=0.1615\textwidth]{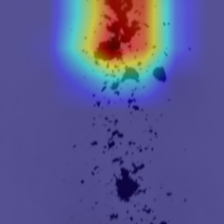}\\
\end{tabularx}

    \caption{\textbf{Saliency visualization.} Grad-CAM maps \cite{Selvaraju:2020:VED} of our TCN with a ResNet-34 backbone. Saliency is mostly located in regions with falling coffee grounds, if present. Color encoding: \colorindicator{sa1}\colorindicator{sa2}\colorindicator{sa3} (low $\to$ high saliency).} %
    \label{fig:gradcam}
\end{figure}

\subsubsection{Generalization.}
All experiments until this point evaluate exactly the same types of coffee beans in the training and test sets. This leaves open whether our proposed methods overfit only to the three types of beans we captured or \emph{generalize} to new types. To gain insights into the generalization capabilities, we train our best-performing model, the TCN with a ResNet-34 backbone, on subsets of our training dataset containing only two types, and we evaluate them on the full test set in \cref{tab:loocv}. We find that the overall accuracy of these models is lower than that of the model trained on the full training set, but this is most likely due to the reduced number of training samples. More interestingly, we find that the \MAED of the unseen bean type is only slightly higher than the \MAED of the types contained in the respective training subset. This demonstrates that our methods can generalize to unseen bean types. Another insight from this experiment is that models trained on bean type \emph{B} are significantly more accurate than those without it. We observe that this type of coffee tends to produce fewer lumps than the other two, and these data seem beneficial for overall training.

\subsubsection{Computational Cost \emph{vs.} Accuracy.}
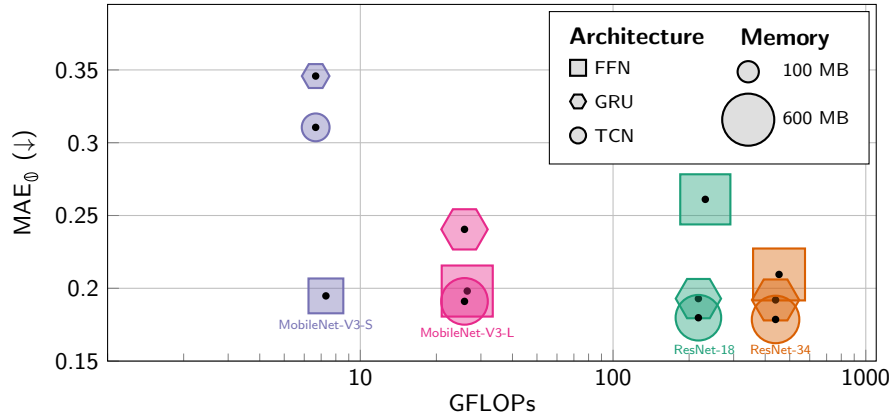
\begin{figure}[t]
    \centering
    
\pgfplotstableread[col sep=comma]{bench_ff.csv}{\dataFF}%
\pgfplotstablegetrowsof{\dataFF}%
\pgfmathtruncatemacro{\rowsFF}{\pgfplotsretval-1}%
\pgfplotstableread[col sep=comma]{bench_gru.csv}{\dataGRU}%
\pgfplotstablegetrowsof{\dataGRU}%
\pgfmathtruncatemacro{\rowsGRU}{\pgfplotsretval-1}%
\pgfplotstableread[col sep=comma]{bench_tcn.csv}{\dataTCN}%
\pgfplotstablegetrowsof{\dataTCN}%
\pgfmathtruncatemacro{\rowsTCN}{\pgfplotsretval-1}%
\tikzset{every picture/.style={/utils/exec={\sffamily\small}}}
\begin{tikzpicture}
    \begin{axis}[
            width=11.75cm,
            height=6.3cm,
            xlabel={GFLOPs},
            ylabel={\MAED ($\downarrow$)},
            xmode=log,
            log ticks with fixed point,
            log basis x=10,
            xlabel shift=-3pt,
            ymajorgrids,
            xmajorgrids, 
            xtick pos=bottom,
            ytick pos=left,
            ticklabel style = {font=\small},
            enlarge x limits=0.0,
            xtick={10, 100, 1000},
            minor xtick={1, 2, 3, 4, 5, 6, 7, 8, 9,
                         20, 30, 40, 50, 60, 70, 80, 90,
                         200, 300, 400, 500, 600, 700, 800, 900
            },
            xticklabels={10, 100, 1000},
            ytick={0.15, 0.2, 0.25, 0.3, 0.35},
            yticklabels={0.15, 0.2, 0.25, 0.3, 0.35},
            ticklabel style = {font=\small},
            ymin=0.15,
            ymax=0.395,
            xmin=1,
            xmax=1100,
            every axis plot/.append style={
                only marks,
                thick
            },
            nodes near coords={\detokenize\expandafter{\pgfplotspointmeta}},
            point meta=explicit symbolic,
            every node near coord/.append style={
                font=\sffamily\tiny,
                yshift=0pt,
                xshift=0pt,
                inner sep=10pt,
                text height=1ex, 
                text depth=0ex,
                text opacity=1   
            },
        ]
        \pgfplotsinvokeforeach{0,...,\rowsFF}{
            \pgfplotstablegetelem{#1}{label_anchor}\of{\dataFF}
            \edef\myanchor{\pgfplotsretval}
            
            \pgfplotstablegetelem{#1}{peak_cuda_memory_mb}\of{\dataFF}
            \pgfmathsetmacro{\bubblesize}{sqrt(\pgfplotsretval) * 0.4}
            
            \pgfmathtruncatemacro{\colorindex}{mod(#1,8)}
            \edef\mycolor{\ifcase\colorindex Dark2-A\or Dark2-B\or Dark2-C\or Dark2-D\or Dark2-E\or Dark2-F\or Dark2-G\or Dark2-H\fi}
            
            \edef\tempaddplot{
                \noexpand\addplot[
                    color=\mycolor, 
                    mark=square*, 
                    mark size=\bubblesize pt,
                    fill opacity=0.4, 
                    draw opacity=1,   
                    nodes near coords style={anchor=\myanchor} 
                ] table [
                    x=gflops,
                    y=full_mae_doppio,
                    restrict expr to domain={\noexpand\coordindex}{#1:#1}
                ] {\noexpand\dataFF};
                
                \noexpand\addplot[
                    color=black,
                    mark=*,
                    mark size=1pt,
                    nodes near coords={}, 
                    forget plot           
                ] table [
                    x=gflops,
                    y=full_mae_doppio,
                    restrict expr to domain={\noexpand\coordindex}{#1:#1}
                ] {\noexpand\dataFF};
            }
            \tempaddplot
        }
        \pgfplotsinvokeforeach{0,...,\rowsGRU}{
            \pgfplotstablegetelem{#1}{label_anchor}\of{\dataGRU}
            \edef\myanchor{\pgfplotsretval}
            
            \pgfplotstablegetelem{#1}{peak_cuda_memory_mb}\of{\dataGRU}
            \pgfmathsetmacro{\bubblesize}{sqrt(\pgfplotsretval) * 0.4}
            
            \pgfmathtruncatemacro{\colorindex}{mod(#1,8)}
            \edef\mycolor{\ifcase\colorindex Dark2-A\or Dark2-B\or Dark2-C\or Dark2-D\or Dark2-E\or Dark2-F\or Dark2-G\or Dark2-H\fi}
            
            \edef\tempaddplot{
                \noexpand\addplot[
                    color=\mycolor,
                    mark=hexagon*, 
                    mark size=\bubblesize pt,
                    fill opacity=0.4,
                    draw opacity=1,
                    nodes near coords style={anchor=\myanchor} 
                ] table [
                    x=gflops,
                    y=full_mae_doppio,
                    restrict expr to domain={\noexpand\coordindex}{#1:#1}
                ] {\noexpand\dataGRU};
                
                \noexpand\addplot[
                    color=black,
                    mark=*,
                    mark size=1pt,
                    nodes near coords={}, 
                    forget plot           
                ] table [
                    x=gflops,
                    y=full_mae_doppio,
                    restrict expr to domain={\noexpand\coordindex}{#1:#1}
                ] {\noexpand\dataGRU};
            }
            \tempaddplot
        }
        \pgfplotsinvokeforeach{0,...,\rowsTCN}{
            \pgfplotstablegetelem{#1}{label_anchor}\of{\dataTCN}
            \edef\myanchor{\pgfplotsretval}
            
            \pgfplotstablegetelem{#1}{peak_cuda_memory_mb}\of{\dataTCN}
            \pgfmathsetmacro{\bubblesize}{sqrt(\pgfplotsretval) * 0.4}
            
            \pgfmathtruncatemacro{\colorindex}{mod(#1,8)}
            \edef\mycolor{\ifcase\colorindex Dark2-A\or Dark2-B\or Dark2-C\or Dark2-D\or Dark2-E\or Dark2-F\or Dark2-G\or Dark2-H\fi}
            
            \edef\tempaddplot{
                \noexpand\addplot[
                    color=\mycolor,
                    mark=*, 
                    mark size=\bubblesize pt,
                    fill opacity=0.4,
                    draw opacity=1,
                    nodes near coords style={anchor=\myanchor} 
                ] table [
                    x=gflops,
                    y=full_mae_doppio,
                    restrict expr to domain={\noexpand\coordindex}{#1:#1}
                ] {\noexpand\dataTCN};
                
                \noexpand\addplot[
                    color=black,
                    mark=*,
                    mark size=1pt,
                    nodes near coords={}, 
                    forget plot           
                ] table [
                    x=gflops,
                    y=full_mae_doppio,
                    restrict expr to domain={\noexpand\coordindex}{#1:#1}
                ] {\noexpand\dataTCN};
            }
            \tempaddplot
        }
        \node[font=\sffamily\tiny, text=Dark2-A, anchor=center] at (230, 0.158) {ResNet-18};
        \node[font=\sffamily\tiny, text=Dark2-B, anchor=center] at (460, 0.158) {ResNet-34};
        \node[font=\sffamily\tiny, text=Dark2-C, anchor=center] at (7.3, 0.175) {MobileNet-V3-S};
        \node[font=\sffamily\tiny, text=Dark2-D, anchor=center] at (26.5, 0.17) {MobileNet-V3-L};
    \end{axis}
    \node[
        draw=black,
        fill=white,
        anchor=north east,
        align=center,
        font=\sffamily\scriptsize,
        inner sep=1.5ex
    ] at ([xshift=-2.5pt, yshift=-2.5pt]current axis.north east) {
        \begin{tabular}{cl}
            \multicolumn{2}{c}{\small\textbf{Architecture}} \\[1ex]
            \tikz[baseline=-0.75ex] \filldraw[fill=gray!25, draw=black, thick] (-3pt,-3pt) rectangle (3pt,3pt); & FFN \\[1.5ex]
            
            \tikz[baseline=-0.75ex] \filldraw[fill=gray!25, draw=black, thick] (3pt,0) -- (1.5pt,2.6pt) -- (-1.5pt,2.6pt) -- (-3pt,0) -- (-1.5pt,-2.6pt) -- (1.5pt,-2.6pt) -- cycle; & GRU \\[1.5ex]
            
            \tikz[baseline=-0.75ex] \filldraw[fill=gray!25, draw=black, thick] (0,0) circle (3pt); & TCN \\
        \end{tabular}%
        \hspace{1pt} %
        \begin{tabular}{cl}%
            \multicolumn{2}{c}{\small\textbf{Memory}} \\[1ex]
            \tikz[baseline=-0.5ex] \draw[black, fill=gray!25, thick] (0,0) circle ({sqrt(100)*0.4 pt}); & 100 MB \\[1.25ex]
            \tikz[baseline=-0.5ex] \draw[black, fill=gray!25, thick] (0,0) circle ({sqrt(600)*0.4 pt}); & 600 MB \\
        \end{tabular}%
    };
\end{tikzpicture}

    \caption{\textbf{Cost-accuracy tradeoffs} of our proposed methods and backbones. We measure the memory and FLOPs for processing a \num{60}-frame sequence (\SI{1}{\second} of video) and report accuracy on the \textit{full} test-set sequence.} %
    \label{fig:complexity}
\end{figure}
To evaluate trade-offs between computational cost and model accuracy, we measure the memory footprint and the number of required floating-point operations (FLOPs) for all proposed architectures on a \num{60}-frame sequence (\SI{1}{\second} of video material). The results of this evaluation are presented in \cref{fig:complexity}.
Our lightweight MobileNet-V3 backbone uses minimal computational power and memory; further, it delivers competitive accuracy with much larger models like ResNet-34 when used as FFN architecture.
However, when we use these small backbones in our GRU- or TCN-based architecture, we see a large decrease in accuracy. This suggests that the highly compressed spatial features extracted by MobileNets lack the capacity needed to build meaningful recurrent states within the GRU and TCN.

In contrast, the ResNet-based FFNs perform worse than their GRU and TCN-based counterparts. Overall, the TCN with a ResNet backbone achieves the highest accuracy across all experiments, but comes at a drastically increased cost in required compute operations, compared to our more lightweight backbones.

\section{Conclusion}
In this paper, we present a case study on contactless weight estimation for falling particles using ground coffee. To enable this research, we captured and introduced \emph{Doppio}, a dataset containing precise, per-frame ground-truth weight measurements of falling coffee grounds at different grind sizes and coffee bean types. We utilized our dataset to conduct a comprehensive evaluation of distinct architectures, including feed-forward networks, GRUs, and TCNs. We benchmarked a diverse range of backbone architectures, ranging from lightweight MobileNet\nobreakdash-V3 variants to ResNets. Our analysis of computational costs highlights the trade-offs between architectural families. We demonstrated that highly compressed models are well-suited for efficient feed-forward processing, but heavier convolutional networks are necessary to build meaningful temporal contexts in GRUs. Further, we demonstrate the generalization capabilities of our model to unseen types of beans. A promising direction for future work is the exploration of quantization-aware training and model quantizations to ensure these models can be efficiently deployed on low-cost, resource-constrained hardware. In conclusion, this work provides an easily accessible dataset and testbed for robust, real-time, vision-based mass estimation of falling particles.

{\subsubsection{Acknowledgments}
SK has been funded by the Deutsche Forschungsgemeinschaft (DFG, German Research Foundation) under Germany's Excellence Strategy --~EXC-3057.
J-MS has been funded by the State of Hesse through LOEWE emergenCITY (Grant no.\ LOEWE/1/12/519/03/05.001(0016)/72).
JG has been funded by the European Research Council (ERC) under the European Union’s Horizon 2020 research and innovation programme (grant agreement No.\ 866008).
Christoph Reich is supported by the Konrad Zuse School of Excellence in Learning and Intelligent Systems (\href{https://eliza.school}{ELIZA}) through the DAAD programme Konrad Zuse Schools of Excellence in Artificial Intelligence, sponsored by the German Federal Ministry of Education and Research. We also acknowledge the support of the European Laboratory for Learning and Intelligent Systems (ELLIS) and Munich Center for Machine Learning (MCML).
SSM and PW have been funded by the DFG --~project No.\ 529680848.
Finally, we thank L. Kammeyer, J. Milkovits, and F. Wichert for their help with recording this dataset.
}

\bibliographystyle{splncs04}
\bibliography{egbib}

\clearpage
\appendix
\renewcommand{\thepage}{\roman{page}}
\setcounter{page}{1}
\renewcommand{\thefigure}{A.\arabic{figure}}
\setcounter{figure}{0}
\renewcommand{\thetable}{A.\arabic{table}} 
\setcounter{table}{0}
\renewcommand{\theequation}{A.\arabic{equation}} 
\setcounter{equation}{0}

\maketitlesupplementary
\thispagestyle{empty}
\label{app:supplement-master}
\FloatBarrier

\begin{table}[h]
    \centering
    \caption{\textbf{Results on different loss weightings.} We report results of our models for $\lambda=10$ and $\lambda=100$ on \emph{Doppio} test, using \MAED and \MAEA (both $\downarrow$) for the \textit{full} sequences. We highlight the best results for each metric \emph{per row} in orange \colorindicator{best}.}
    
\footnotesize%
\setlength{\tabcolsep}{3.0pt}%
\renewcommand{\arraystretch}{0.925}
\hspace{-3.3pt}\begin{tabularx}{\figureWidth}{c@{\hspace{5pt}} | @{\hspace{4pt}} X S[table-format=1.2,round-mode=places,round-precision=2] S[table-format=1.2,round-mode=places,round-precision=2] l S[table-format=1.2,round-mode=places,round-precision=2] S[table-format=1.2,round-mode=places,round-precision=2]}
        \toprule
        \multicolumn{1}{c}{} & \multirow{2}{*}[-0.5ex]{\textbf{Backbone}} & \multicolumn{2}{c}{$\lambda = 10$} & \hphantom{04} &\multicolumn{2}{c}{$\lambda = 100$}\\
        \cmidrule(lr{14pt}){3-4} \cmidrule(r){6-7}
        \multicolumn{1}{c}{} & & {\textbf{\MAED}} & {\textbf{\MAEA}} & & {\textbf{\MAED}} & {\textbf{\MAEA}}\\
        \midrule
        
        \multirow{4}{*}{\rotatebox[origin=c]{90}{\textbf{FFN}}}
        & MobileNet-V3-S & \bestcell0.1948 & 0.2523 & & 0.3299 & \bestcell0.2351\\
        & MobileNet-V3-L & \bestcell0.1981 & 0.2602 & & 0.3429 & \bestcell0.2520\\
        & ResNet-18 & \bestcell0.2611 & 0.3003 & & 0.2986 & \bestcell0.2113\\
        & ResNet-34 & \bestcell0.2095 & 0.2411 & & 0.2645 & \bestcell0.2043\\
        \midrule
        
        \multirow{4}{*}{\rotatebox[origin=c]{90}{\textbf{GRU}}} 
        & MobileNet-V3-S & 0.3736 & 0.3426 & & \bestcell0.3458 & \bestcell0.3249\\
        & MobileNet-V3-L & 0.2596 & \bestcell0.2279 & & \bestcell0.2405 & 0.2588\\
        & ResNet-18 & 0.1996 & \bestcell0.2427 & & \bestcell0.1929 & 0.2616\\
        & ResNet-34 & 0.2399 & \bestcell0.1952 & & \bestcell0.1920 & 0.2555\\
        \midrule

        \multirow{4}{*}{\rotatebox[origin=c]{90}{\textbf{TCN}}}
         & MobileNet-V3-S & \bestcell0.3114 & \bestcell0.2607 & & \bestcell0.3105 & 0.3723\\
        & MobileNet-V3-L & 0.254 & \bestcell0.217 & &\bestcell0.1910 & 0.2274\\
         & ResNet-18 & \bestcell0.1752 & \bestcell0.2337 & & \bestcell0.1798 & \bestcell0.2321\\
        & ResNet-34 & 0.2232 & 0.2399 & & \bestcell0.1786 & \bestcell0.2325\\
        \bottomrule
    \end{tabularx}

    \label{fig:suppl-lambda}
\end{table}
\section{Loss Weighting}

\noindent As described in Sec. 5.2, we provide the full evaluation of the loss weighting factor $\lambda$ for the different architecture-types in \cref{fig:suppl-lambda}. As reported in the main paper, we find that $\lambda=\text{\num{10}}$ performs best for \MAED of the feed-forward network, and the differences between the \MAEA in both settings are rather small. At the same time, we find $\lambda=\text{\num{100}}$ to be the best for GRU- and TCN-based architectures in terms of \MAED, and the \MAEA is only slightly smaller for $\lambda=10$. 
\newpage
\begin{table}[h]
    \centering
    \caption{\textbf{Results on different distance measures.} We report results of our models trained with different distance measures within their respective loss functions on \emph{Doppio} test, using \MAED and \MAEA (both $\downarrow$) for the \textit{full} sequences. We highlight the best results for each metric \emph{per row} in orange \colorindicator{best}.}
    
\footnotesize%
\setlength{\tabcolsep}{3.0pt}%
\renewcommand{\arraystretch}{0.925}
\hspace{-3.3pt}\begin{tabularx}{\figureWidth}{c@{\hspace{5pt}} | @{\hspace{4pt}} X S[table-format=1.2,round-mode=places,round-precision=2] S[table-format=1.2,round-mode=places,round-precision=2] l S[table-format=1.2,round-mode=places,round-precision=2] S[table-format=1.2,round-mode=places,round-precision=2] l S[table-format=1.2,round-mode=places,round-precision=2] S[table-format=1.2,round-mode=places,round-precision=2]}
        \toprule
        \multicolumn{1}{c}{} & \multirow{2}{*}[-0.5ex]{\textbf{Backbone}} & \multicolumn{2}{c}{Smooth L1} & \hphantom{04} &\multicolumn{2}{c}{MAE} & & \multicolumn{2}{c}{MSE}\\
        \cmidrule(lr{14pt}){3-4} \cmidrule(r){6-7} \cmidrule(r){9-10}
        \multicolumn{1}{c}{} & & {\textbf{\MAED}} & {\textbf{\MAEA}} & & {\textbf{\MAED}} & {\textbf{\MAEA}} & & {\textbf{\MAED}} & {\textbf{\MAEA}}\\
        \midrule
        
        \multirow{4}{*}{\rotatebox[origin=c]{90}{\textbf{FFN}}}
        & MobileNet-V3-S & \bestcell0.1948 & \bestcell0.2523 && 0.2194 & 0.2565 && 0.1979 & 0.2635\\
        & MobileNet-V3-L & \bestcell0.1981 & 0.2602 && 0.2594 & 0.1962 && \bestcell0.1984 & \bestcell0.2278\\
        & ResNet-18 & \bestcell0.2611 & 0.3003 && 0.2654 & \bestcell0.2520 && 0.324 & 0.3054\\
        & ResNet-34 & \bestcell0.2095 & \bestcell0.2411 && 0.2540 & \bestcell0.2441 && 0.2279 & 0.2963 \\
        \bottomrule
    \end{tabularx}

    \label{fig:suppl-dist}
\end{table}
\section{Distance Measures}
\noindent Our loss terms defined in Eq. (3) and Eq. (4) require a distance measure. We provide the evaluation of different distance functions in \cref{fig:suppl-dist}. As we reported in the main paper, we find that smooth L1 performs best in our case. MAE-based distance functions outperforming MSE-based ones in our case can be explained by the fact that the predicted weights per step are all below \SI{1}{\gram}, so the quadratic penalty actually reduces their magnitude.

\end{document}